# TOWARDS THE DEVELOPMENT OF A SIMULATOR FOR INVESTIGATING THE IMPACT OF PEOPLE MANAGEMENT PRACTICES ON RETAIL PERFORMANCE

Peer-Olaf Siebers[1], Uwe Aickelin[1], Helen Celia[2], Chris W. Clegg[2]

[1]University of Nottingham, School of Computer Science (IMA)
Nottingham, NG8 1BB, UK

[2]University of Leeds, Centre for Organisational Strategy, Learning & Change, LUBS
Leeds, LS2 9JT, UK

**ABSTRACT**
Often models for understanding the impact of management practices on retail performance are developed under the assumption of stability, equilibrium and linearity, whereas retail operations are considered in reality to be dynamic, non-linear and complex. Alternatively, discrete event and agent-based modelling are approaches that allow the development of simulation models of heterogeneous non-equilibrium systems for testing out different scenarios.

When developing simulation models one has to abstract and simplify from the real world, which means that one has to try and capture the 'essence' of the system required for developing a representation of the mechanisms that drive the progression in the real system. Simulation models can be developed at different levels of abstraction. To know the appropriate level of abstraction for a specific application is often more of an art than a science. We have developed a retail branch simulation model to investigate which level of model accuracy is required for such a model to obtain meaningful results for practitioners.

**1 INTRODUCTION**
The retail sector has been identified as one of the biggest contributors to the productivity gap that persists between the UK and other countries, in particular France, Germany and the USA (Reynolds *et al*, 2005). UK retail productivity measures paint a worrying picture, describing lower levels of efficiency than what we would expect (Department of Trade and Industry, 2003), and in particular lower than the benchmark countries already stated. Researchers have so far failed to explain fully the reasons accounting for the productivity gap, and management practices provide a valid and thoughtful way of looking at the problem. The analysis of management practices across different contexts has attempted to explain differences in organisational productivity and performance (for a review see Wall and Wood, 2005).

A recent report on UK productivity asserted that "... the key to productivity remains what happens inside the firm and this is something of a 'black box'…" (Delbridge *et al*, 2006). Siebers and colleagues conducted a comprehensive literature review of this research area to assess linkages between management practices and organisational productivity (Siebers *et al*, 2008). The authors concluded that management practices are multidimensional constructs that generally do not demonstrate a straightforward relationship with productivity variables. Empirical evidence affirms that both management practices and productivity measures must be context specific to be effective. Management practices need to be tailored to the particular organisation and the working environment, whereas productivity indices must also reflect a particular organisation's activities on a local level to be a valid indicator of performance.

It is challenging work to try and delineate the effects of management practices from other socially embedded factors. Most Operations Research (OR) methods can be applied as analytical tools once management practices have been implemented, however they are not very useful at revealing system-level effects prior to the introduction of specific management practices. This is most restricting when the focal interest is the development of the system over time, because dynamic behaviour is a fundamental feature of many interesting real-world phenomena. An alternative to these methods, which offers potential to overcome such limitations, is simulation. In particular Agent-Based Simulation (ABS) seems to be well suited to the investigation of people centric complex adaptive systems, of which a retail department is an example.

There has been a lot of modelling and simulation of operational management practices, but people management practices, for example training, empowerment and teamwork, have often been neglected. This seems a fertile area for research as recent findings suggest that people management practices significantly impact on a business' productivity (Birdi *et al*, 2008). One reason for the paucity of modelling and simulation of people management practices relates to their key component, an organisation's people, who are often unpredictable in their individual behaviour.

Our overall aim is to investigate if we can understand the relationship between people management practices and retail performance better by employing simulation as the technology for evaluating different strategies, considering that the system of interest is dynamic, non-linear and complex. In pursuit of this aim, we studied the following two research questions: (1) Is an event driven simulation a suitable tool to better understand the relationship between people management practices and retail performance? (2) What level of abstraction should we use for our simulation model?

To answer these research questions we have adopted a case study approach using applied research methods to collect both qualitative and quantitative data. We have worked with a leading UK retail organisation where we have conducted some surveys and four weeks of informal participant observation in four departments across two retail branches. The approach has enabled us to acquire a valid and reliable understanding of how the real system operates, revealing insights into the working of the system as well as the behaviour of and interactions between the different individuals and their complementary roles within the retail department. Using the obtained knowledge and the data, we have developed conceptual models of the real system and the individuals within the system and implemented these in a simulation model.

The simulation model has been developed in two major steps, each related to one of the research questions defined above. First, we have focused on putting together a functional representation of the real system using a mixed process oriented Discrete Event Modelling (DEM) and individual oriented Agent-Based Modelling (ABM) approach and we have tested if such an approach is of use for investigating the impact of people management practices on productivity and customer satisfaction. After obtaining a positive answer to the first research question (Siebers *et al*, 2009) in this paper we are focusing on investigating the second research question, i.e. the level of model accuracy required to obtain some meaningful results for practitioners. For this purpose, we have added some more details to the model (in form of additional algorithms and empirical data) and tested the sensitivity of the simulation model towards these additions. We have also compared the system performance predicted by the simulation model (without using any form of calibration) to the system performance measured in the real system, in order to access the predictive capabilities (qualitative or quantitative) of our simulation model in its current form.

The investigations described in this paper help practitioners to decide how to best coordinate time and effort when having limited resources for conducting similar simulation studies.

In Section 2 we give an overview of the relevant literature, looking at the different modelling approaches used in OR and at previous attempts to model people centric systems in an OR context. Section 3 describes the model design and its implementation. Section 4 presents two experiments we have conducted to validate the final simulation model and to test the sensitivity of the model output towards some of the features we have added in the second development step of our simulation model. Section 5 concludes the paper with a summary and an outlook of further developments.

**2 BACKGROUND**

OR is applied to problems concerning the conduct and co-ordination of the operations within an organisation (Hillier and Lieberman, 2005). An OR study usually involves the development of a scientific model that attempts to abstract the essence of the real problem. When investigating the behaviour of complex systems the choice of an appropriate modelling technique is very important. To inform the choice of an appropriate modelling technique we have reviewed relevant literature from the fields of Economics, Social Sciences, Psychology, Retail Management, Supply Chain Management, Manufacturing, Marketing, OR, Artificial Intelligence, and Computer Science.

**2.1 MODELLING APPROACHES**
Within the fields listed above a wide variety of approaches have been used which can be broadly classified into three categories: analytical approaches, heuristic approaches and simulation. Often we found that combinations of these were used within a single model (e.g. Schwaiger and Stahmer, 2003; Greasley, 2005).

*2.1.1 Analytical methods*
Once data has been collected, it is common in economics, Social Science, and Psychology to use analytical analysis tools to quantify causal relationships between different factors. Often some form of regression analysis is used to investigate the correlation between independent and dependent variables. Patel and Schlijper (2004) use a multitude of different analytical and other modelling approaches to test specific hypotheses of consumer behaviour. Another good example of this type of analysis can be found in Clegg *et al* (2002) who investigate the use and effectiveness of modern manufacturing practices. The survey data is analysed using parametric and nonparametric analytical techniques, as appropriate to the nature of the response scales and the distributions of scores obtained.

*2.1.2 Heuristic methods*
No relevant purely heuristic models were found during the literature review. This does not come as a surprise as pure heuristic models are more frequently used in system optimisation, i.e. not the focus of our current research.

*2.1.3 Simulation*
Simulation introduces the possibility of a new way of thinking about social and economic processes based on ideas about the emergence of complex behaviour from relatively simple activities (Simon, 1996). It allows clarification of a theory and investigation of its implications.

OR usually employs three different types of simulation modelling to help understand the behaviour of organisational systems, each of which has its distinct application area: System Dynamics (SD), Discrete Event Simulation (DES) and ABS. SD takes a top down approach by modelling system changes over time. The analyst has to identify the key state variables that define the behaviour of the system and these are then related to each other through coupled, differential equations. SD is applied where individuals within the system do not have to be highly differentiated and knowledge on the aggregate level is available, for example modelling population, ecological and economic systems.

DES is a process centric modelling approach, i.e. the focus is on process flow. The system is modelled as a set of entities (passive objects) that are processed and evolve over time according to the availability of resources and the triggering of events (Law and Kelton, 1991). The simulation model maintains an ordered queue of events. DES is commonly used as a decision support tool in the manufacturing and service industries.

ABS is well suited for modelling systems with a heterogeneous, autonomous and proactive population, and is therefore well suited to analyse people centric systems. It is a bottom-up approach where the modeller has to identify the active entities in the system, defines their behaviours, puts them in an environment, and establishes their connections. Macro behaviour at the system level does not have to be modelled explicitly; it emerges as a result from the interactions between the individual entities, also called agents (Pourdehnad *et al*, 2002). These agents are autonomous discrete entities with individual goals and behaviours, where autonomy refers to the fact that they act independently and are not guided by some central control authority or process. (Bakken, 2007). In addition, agents are capable of behaving proactively, i.e. initiating changes rather than just reacting to events.

ABS is suited to a system driven by interactions between its constituent entities, and can reveal what appears to be complex emergent behaviour at the system level even when the agents involved exhibit fairly simple behaviours on a micro-level. Some typical application domains for ABS are ecology, traffic and transportation, sociology, economic system analysis, and gaming. Out of the simulation approaches reviewed ABS seems to be the most suitable one for our purpose due to it autonomous entity focus.

**2.2 COMPARING APPROACHES**
Most methods to study managerial implications can only be applied as analytical tools once management practices have been implemented, however they are not very useful at revealing system-level effects prior to the introduction of specific management practices. Simulation is a what-if analysis tool that allows to study different management practices scenarios prior to their implementation without interrupting the operation of the real system. Furthermore, often analytical models are developed under the assumption of stability, equilibrium and linearity. However, retail operations are considered in reality to be dynamic, non-linear and complex. Simulation allows developing models of non-equilibrium systems at any level of complexity.

While analytical models typically aim to explain correlations between variables measured at a single point in time, simulation models are concerned with the development of a system over time (Law and Kelton, 1991). Therefore, simulation provides an insight into system dynamics rather than just predicting the output of a system based on specific inputs. In addition, it allows visualising the changes of key variables over time, which provides useful information about the dynamic behaviour of the system.

ABS in particular has some additional features that are very useful for modelling people centric systems. It supports the understanding of how the dynamics of real systems arise from the characteristics of individuals and their environment. It allows modelling a highly diverse population where each agent can have personal motivations and incentives, and to represent groups and group interactions. Furthermore, human behaviour is often quite irrational. For example, fleeing a fire people will often try to retrace their steps and leave the building by the way they came in, rather than heading for the nearest exit - even if it is much closer. While other simulation modelling techniques do not allow the consideration of such concepts ABS allows representing irrational behaviour. Another advantage of ABS is that building an ABS model is a very intuitive process as actors in the real system can be directly modelled as agents in the simulation model. This also supports the micro validation (i.e. the validation of the agent templates) of the simulation model as it is easy for an expert in the real system (who is not an expert in ABS) to quickly take on board the model conceptualisation and provide useful validation of the model component structures and content.

Nevertheless, there are also some disadvantages associated with ABS. Often people argue about computational resources, stating that ABS requires massive resources compared to other simulation technologies (e.g. Rahmandad and Sterman, 2008). However, we have made the experience in previous projects that with today's technology in this particular area of application (where the number of agents simulated at the same time is relatively small) simulation run times are acceptable when using Java as a programming language. There also seems consensus in the literature that it is difficult to empirically evaluate agent-based models, in particular at the macro level, because the behaviour of the system emerges from the interactions between the individual entities (Moss and Edmonds, 2005). Furthermore, problems often occur through a lack of adequate empirical data; it has been questioned whether a model can be considered a scientific representation of a system when it has not been built with 100% objective, measurable data. However, many of the variables built into a system cannot be quantified objectively. In such cases, expertly-validated estimates offer a unique solution to the problem. Finally, Twomey and Cadman (2002) state that there is always a danger that people new to ABS may expect too much from an ABM, in particular with respect to predictive ability (a caveat which in fact applies to all the simulation approaches mentioned above). To mitigate this problem it is important to be clear with individuals about what this modelling technique can really offer in order to guide realistic expectations.

In conclusion, we can say that simulation, and in particular ABS, offers a fresh opportunity to realistically and validly model organisational characters and their interactions, which in turn can facilitate a meaningful investigation of management practices and their impact on system outcomes.

### 2.3 RELATED RESEARCH
We have found that most of the work relevant to our investigations focuses on marketing and consumer behaviour rather then on management practices. For example, Said *et al* (2002) have created an ABS composed of a virtual consumer population to study the effects of marketing strategies in a competing market context. A similar approach has been used by Baxter et al (2003) who have developed an intelligent customer relationship management tool using ABS that considers the communication of customer experiences between members of a social network, incorporating the powerful influence of word of mouth on the adoption of products and services. A very different facet of consumer behaviour is presented by Kitazawa and Batty (2004) who investigate the retail movements of shoppers in a large shopping centre. There are many more examples where ABM has been employed to study consumer behaviours (e.g. Cao, 1999; Csik, 2003; Jager *et al*, 2000; Baydar, 2003) or entire consumer market behaviours (e.g. Vriend, 1995; Twomey and Cadman, 2002; Said and Bouron, 2001; Koritarov, 2004; Janssen and Jager, 2001; Schenk *et al*, 2007).

While most of the relevant papers reviewed apply ABM and ABS there are some noteworthy exceptions. For example, Berman and Larson (2004) use queue control models to investigate the efficiency of cross-trained workers in stores. Another interesting contribution is made by Nicholson *et al* (2002), who compare different marketing strategies for multi channel (physical and electronic) retailing, applying a traditional Belkian analysis of situational variables in a longitudinal study of consumer channel selection decisions. As a last example, we want to mention Patel and Schlijper (2004) who use a multitude of different analytical and other modelling approaches to test specific hypotheses of consumer behaviour.

Finally, it is worthwhile mentioning that we have found one of-the-shelf software, ShopSim (http://www.savannah-simulations.com/ accessed 30/07/2009), which is a decision support tool for retail and shopping centre management. It evaluates the shop mix attractiveness and pedestrian friendliness of a shopping centre. The software uses an agent-based approach, where the behaviour of agents depends on poll data.

### 2.4 OUR CHOICE
When modelling people management practices in OR one is mainly interested in the relations between staff and customers but it is equally important to consider parts of the operational structure of the system in order to enable a realistic evaluation of system performance and customer satisfaction. Therefore, we have decided to use a mixed DEM and ABM approach for our simulation model. A queuing system will be used to model the operational structure of the system while the people within the system will be modelled as autonomous entities (agents) in order to account for the stochasticity caused by human decision making. This will also allow us to consider the long term effects of service quality on the decision making processes of customers, both important components when designing a tool for studying people centric systems. The application of a mixed DEM and ABM approach has been proven to be quite successful in areas like manufacturing and supply chain modelling (e.g. Parunak *et al*, 1998; Darley *et al*, 2004) as well as in the area of modelling crowd movement (e.g. Dubiel and Tsimhoni, 2005).

Finally, we would like to note that our goal is to develop a simulation model that is a genuinely practical model, which incorporates enough realism to yield results that will be of direct value to managers. This differentiates it from many idealised and simplified simulation models in the academic literature. In particular in ABM, although most models have been inspired by observations of real social systems they have not been tested rigorously using empirical data and most efforts do not go beyond a "proof of concept" (Janssen and Ostrom, 2006).

## 3 MODEL

When modelling people inside a system it is important to consider that there are differences in the way ABM is applied in different research fields regarding their empirical embeddednes. Boero and Squazzoni (2005) distinguish between three different levels, amongst others characterised by the level of empirical data used for input modelling and model validation: case-based models (for studying specific circumscribed empirical phenomena), typifications (for studying specific classes of empirical phenomena) and theoretical abstractions (pure theoretical models). While case based models use empirical data for input modelling as well as model validation theoretical abstractions use no empirical data at all. Social Science simulation applications tend to be more oriented towards the bottom end of this scale (theoretical abstractions) OR applications are usually located at the top end (case based). This implies that there is also a difference in knowledge representation and in the outcome that the researcher is interested in. In Social Sciences it is common to model the decision making process itself (e.g. Rao and Georgeff, 1995) and the focus of attention on the output side is on the emergence of patterns. On the other hand in OR applications the decision making process is often represented through probabilities or simple if-then decision rules collected from an existing real system (e.g. Schwaiger, 2007; Darley *et al*, 2004) and the focus on the output side is on system performance rather than on emergent phenomena. As we are studying a people centric service system, we have added some additional measures to assess how people perceive the services provided, besides the standard system performance measures.

We start this section by describing the case studies we have conducted to better understand the problem domain and to gather some empirical data. What we have learned during those case studies is reflected in the conceptual models presented. We have used these together with the empirical data collected as a basis for our implementation, which we describe towards the end of this section. Throughout the rest of this paper we will use the term 'actor' to refer to a person in the real system, whereas the term 'agent' will be reserved for their counterparts in the simulation model. Furthermore, we will use the abbreviation 'ManPraSim' when referring to our management practice simulation model and v1 for the simulation model developed to answer the first research question and v2 for the simulation model developed to answer the second research question.

### 3.1 KNOWLEDGE GATHERING

Case studies were undertaken in four departments across two branches of a leading UK retailer. The case study work involved extensive data collection techniques, spanning: participant observation, semi-structured interviews with team members, management and personnel, completion of survey questionnaires on the effectiveness of retail management practices and the analysis of company data and reports. Research findings were consolidated and fed back (via report and presentation) to employees with extensive experience and knowledge of the four departments in order to validate our understanding and conclusions. This approach has enabled us to acquire a valid and reliable understanding of how the real system operates, revealing insights into the working of the system as well as the behaviour of and interactions between the different actors within it.

In order to make sure that our results regarding the application of management practices are applicable for a wide variety of departments, we have chosen two different types of case study departments which are substantially different not only in their way of operating but also their customer base and staffing setup. We collected our data in the Audio & Television (A&TV) and the WomensWear (WW) departments of the two case study branches.

The two departments can be characterised as follows:
- A&TV: average customer service times is much longer; average purchase is much more expensive; likelihood of customers seeking help is much higher; likelihood of customers making a purchase after receiving help is lower; conversion rate (likelihood of customers making a purchase) is lower; department tends to attract more solution demanders and service seekers (the terminology will be explained in Section 3.4.2)
- WW: average customer service times is much shorter; average purchase is much less expensive; likelihood of customers seeking help is much lower; likelihood of customers making a purchase after receiving is much higher; conversion rate is higher; department tends to attract shopping enthusiasts

### 3.2 CONCEPTUAL MODELLING

Based on the results of our assessment of alternative modelling techniques in the background section and from what we have learned from our case studies we have designed conceptual models of the system to be investigated and the actors within the system.

### 3.2.1 Concept for the simulation model

The initial idea for our ManPraSim and its components is shown in Figure 1. Regarding system inputs we use different types of agents (customers, sales staff and managers), each with a different set of relevant attributes. Furthermore, we define some global parameters that can influence any aspect of the system. The core of our simulation model is a dynamic system representation including a visualisation of system and agent states to allow monitoring the interactions of the agents as well as the system performance at runtime. The simulation model also includes a user interface, which enables some form of user interaction (change of parameters) before and during the runtime. On the output side, we might be able to observe some emergent behaviour on the macro level although this is not our primary objective. What we are mainly interested in are the standard system performances measures like transactions, staff utilisation and some measure of customer satisfaction. Furthermore, we want to use the simulation output to identify bottlenecks in the system and therefore assist with optimisation of the modelled system.

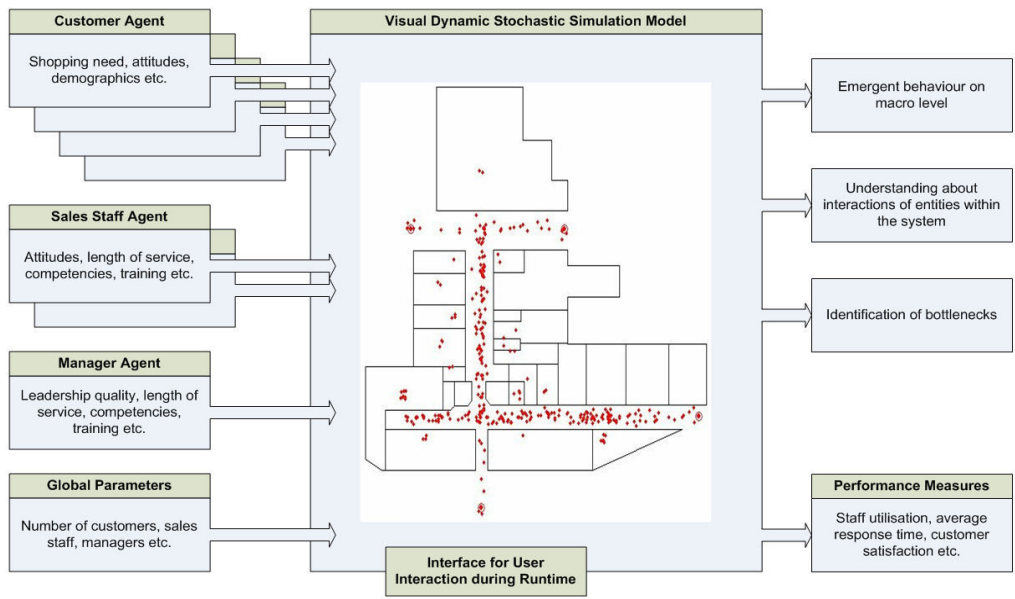

**Figure 1:** Conceptual model of the retail department simulation model.

### 3.2.2 Concepts for the agents

We have used state charts for the conceptual design of our agents. State charts show the different states an entity can be in and define the events that cause a transition from one state to another. This is exactly the information we need in order to represent our agents at a later stage within the simulation environment. We have found this graphical representation a useful part of the agent design process because it is easier for an expert in the real system (who is not an expert in ABM) to quickly take on board the model conceptualisation and provide useful validation of the model component structures and content.

Designing and building a model is to some extent subjective, and the modeller has to selectively simplify and abstract from the real scenario to create a useful model (Shannon, 1975). A model is always a restricted copy of the real world, and an effective model consists of only the most important components of the real system. In our case, our case studies indicated that the key system components take the form of the behaviours of an actor and the triggers that initiate a change from one behavioural state to another. We have developed state charts for all of the agents in our retail department simulation model. Figure 2 shows as an example the conceptual model of a customer agent. Here the transition rules have been omitted to keep the chart comprehensible.

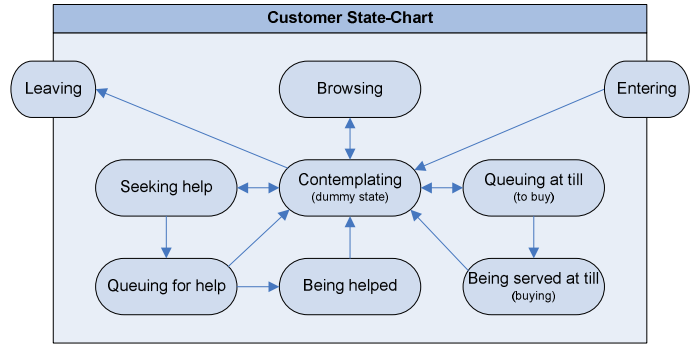

**Figure 2:** Conceptual model of our customer agents.

An important remark about the order in which a customer is going through the different states within the customer state chart is that there is a logical order to events. Some of this order has been expressed with single- and double-headed arrows whereas others would have been difficult to express in the graph directly without losing the concept of the connecting contemplating state. For example, a customer would not be queuing at the till to buy something without having purchased an item. Therefore, the preceding event for a customer queuing at the till is to make a purchase, which in turn requires that the customer has been browsing to pick up an item to purchase. These rules have been considered later in the implementation (see Section 3.4.1 and Figure 3 for more details).

### 3.2.3 Concept for measure of service perception
We have introduced a customer satisfaction level index as a novel performance measure using satisfaction weightings. This new measure is required because existing indices such as queuing times or service times are less useful in modelling services than manufacturing activities. In essence, purely quantitative measures fail to capture the quality of service, which is arguably the most important when considering productivity in retail. The inevitable trade-off between quality and quantity is particularly salient when customers and staff come face to face and therefore we consider this measure of quality in conjunction with others of quantity.

Historically customer satisfaction has been defined and measured in terms of customer satisfaction with a purchased product (Yi, 1990). The development of more sophisticated measures has moved on to incorporate customers' evaluations of the overall relationship with the retail organisation, and a key part of this is the service interaction. Indeed, empirical evidence suggests that quality is more important for customer satisfaction than price or value-for-money (Fornell *et al*, 1996), and extensive anecdotal evidence indicates that customer-staff service interactions are an important determinant of quality as perceived by the customer.

Our customer satisfaction level index allows customer service satisfaction to be recorded throughout the simulated lifetime. The idea is that certain situations might have a bigger impact on customer satisfaction than others, and therefore weights can be assigned to events to account for this. Depending on type of service requested and level of service provided different results turn up. Applied in conjunction with an ABM approach, we expect to observe interactions with individual customer differences; variations which have been empirically linked to differences in customer satisfaction (e.g. Simon and Usunier, 2007). This helps the practitioner to find out to what extent customers underwent a positive or negative shopping experience and it allows the practitioner to put emphasis on different operational aspects and try out the impact of different strategies.

**3.3 EMPIRICAL DATA**

Often agent logic is based on analytical models or heuristics and in the absence of adequate empirical data theoretical models are used. However, as we have explained in Section 3, we have taken a different approach. We have used frequency distributions for determining state change delays and probability distributions for representing the decisions made as statistical distributions are the best format to represent the data we have gathered during our case study due to their numerical nature. In this way, a population is created with individual differences between agents, mirroring the variability of attitudes and behaviours of their real human counterparts.

Our frequency distributions are modelled as triangular distributions supplying the time that an event lasts, using the minimum, mode and maximum duration. We have chosen triangular distributions here as we have only a relatively small sample of empirical data and a triangular distribution is commonly used as a first approximation for the real distribution (XJTEK, 2005). The values for our triangular distributions are based on our own observation and expert estimates in the absence of numerical data. We have collected this information from the two branches and calculated an average value for each department type, creating one set of data for A&TV and one set for WW. Table 1 lists some sample frequency distributions that we have used for modelling the A&TV department (the values presented here are slightly amended to comply with confidentiality restrictions). The distributions are used as exit rules for most of the states. All remaining exit rules are based on queue development, i.e. the availability of staff.

| situation | min | mode | max |
|---|---|---|---|
| leave browse state after … | 1 | 7 | 15 |
| leave help state after … | 3 | 15 | 30 |
| leave pay queue (no patience) after … | 5 | 12 | 20 |

**Table 1:** Sample frequency distribution values.

The probability distributions are partly based on company data (e.g. conversion rates, i.e. the percentage of customers who buy something) and partly on informed guesses (e.g. patience of customers before they would leave a queue). As before, we have calculated average values for each department type. Some examples for probability distributions we used to model the A&TV department can be found in Table 2. The distributions make up most of the transition rules at the branches where decisions are made with what action to perceive (e.g. decision to seek help). The remaining decisions are based on the state of the environment (e.g. leaving the queue, if the queue does not get shorter quickly enough).

| event | probability of event |
|---|---|
| someone makes a purchase after browsing | 0.37 |
| someone requires help | 0.38 |
| someone makes a purchase after getting help | 0.56 |

**Table 2:** Sample probabilities.

We have also gathered some company data about work team numbers and work team composition, varying opening hours and peak times, along with other operational and financial details (e.g. transaction numbers and values).

**3.4 IMPLEMENTATION**

Our conceptual models have been implemented in AnyLogic™ v5.5. This is a Java™ based multi-paradigm simulation software (XJTEK, 2005) which supports the development of mixed DE and AB models. It allows replacing passive objects from the DE model with active objects (agents), which in our case represent the actors of the real world system.

As mentioned in Section 1, the simulation model has been developed in two major steps, each related to one of the research questions. In a first step, we have developed a relatively simple functional representation of the real system. In this simulation model the agents are largely homogeneous and without memory. Therefore, it is impossible to study the long term effects of people management practices on individual customer's satisfaction. However, this simulation model turned out to be useful for studying certain aspects of branch operations and to carry out some experiments for investigating the impact of different people management practices on a strategic level. In a second step, we have added some more features to the simulation model. We have created a finite population of heterogeneous agents with memory and the capability to evolve over time and we have added more accuracy to our department representation by adding empirical footfall data, opening hours varying by day and a special customer egress procedure for the time when the store is closing. The simulation model has been validated and we have conducted a sensitivity analysis. In this section, we describe the functionality and features of the latest version of our simulation model, ManPraSim v2.

### *3.4.1 The main concept*
During the implementation, we have applied the knowledge, experience and data accumulated through our case study work. The simulation model presented here is capable of representing customers, service staff (with different levels of expertise) and managers. Figure 3 shows a screenshot of our customer and staff agent templates in AnyLogic™. Boxes represent states, arrows transitions, arrows with a dot on top entry points, circles with a B inside branches, and the numbers represent satisfaction weights. Service staff and managers are using the same template, only their responsibilities and priorities differ. The system (i.e. the department) is implemented as an array of queues, each of which is served by dedicated staff members.

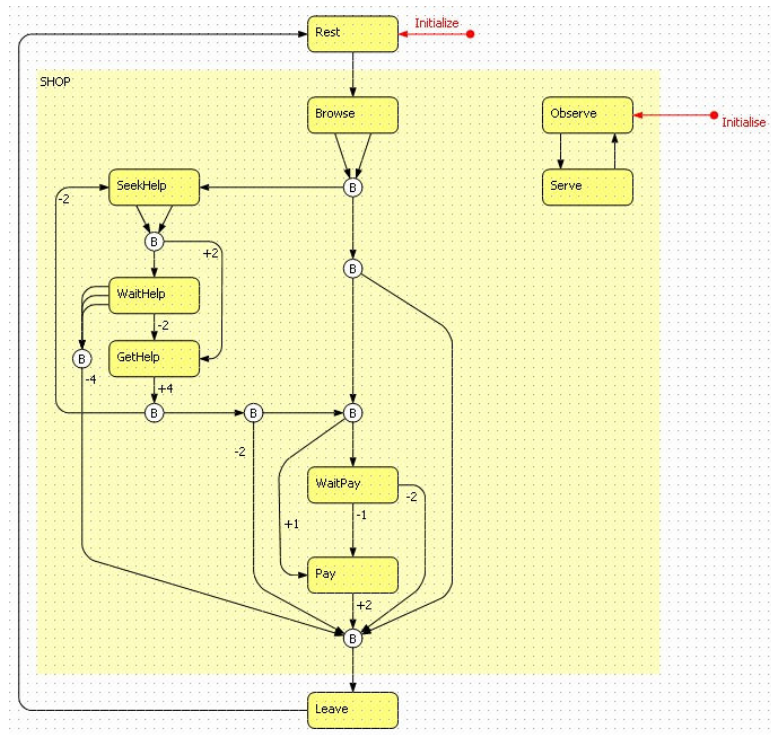

**Figure 3:** Customer (left) and staff (top right) agent logic implementation in AnyLogic™.

At the beginning of each simulation run, a customer pool is created which represents a population of potential customers who can visit the simulated department on an unspecified number of occasions. Once the simulation has started, customers are chosen at a specified rate (customer arrival rate) and released into the simulated department to shop. The customer agent template consists of three main blocks (help and pay) which use a very similar logic. In each block, in the first instance, customers try to obtain service directly from a staff member and if they cannot obtain it (i.e. no suitable staff member is available) they have to queue. They are then served as soon as the suitable staff member becomes available, or leave the queue if they do not want to wait any longer (an autonomous decision). Once customers have finished their shopping (either successfully or not) they leave the simulated department and are added back to the customer pool where they rest until they are picked the next time.

*3.4.2 Special features*
There are certain peak times where the pressure on staff members is higher, which puts them under higher work demands and results in different service times. There is a weekly demand cycle. For example on a Saturday, a lot more customers visit the branch compared to the average weekday. In ManPraSim v2 we have incorporated these real temporal fluctuations in customer arrival rates, across daily variations to opening hours. It includes the calculated hourly footfall values for each of the four case study departments for each hour of the day and each day of the week, based on sales transaction data that is automatically recorded by the company. Conversion rates are based on staff estimates and data from a leading UK retail database. The gaps between customer arrivals are based on exponential distributions, which account for further variation in weekly footfall.

In real life, customers display certain shopping behaviours that can be categorised. Hence, we introduced customer types to ManPraSim v2 to create a heterogeneous customer base, thereby allowing us to test customer populations closer to reality. We have introduced five customer types: shopping enthusiasts, solution demanders, service seekers, disinterested shoppers and internet shoppers. The latter are customers that only seek advice but are likely to buy only from the cheapest source, e.g. the Internet. The first three customer types have been identified by the case study organisation as the customers who make the biggest contribution to their business, in terms of both value and frequency of sales. In order to avoid over-inflating the amount of sales that we model we have introduced two additional customer types, which use services but often do not make a purchase. The definition of each customer type is based on the customer's likelihood to perform a certain action, classified as either: low, moderate, or high. The definitions can be found in Table 3.

| Customer type | Likelihood to | | | |
|---|---|---|---|---|
| | buy | wait | ask for help | ask for refund |
| Shopping enthusiast | high | moderate | moderate | low |
| Solution demander | high | low | low | low |
| Service seeker | moderate | high | high | low |
| Disinterested shopper | low | low | low | high |
| Internet shopper | low | high | high | low |

**Table 3:** Definitions for each type of customer.

A moderate likelihood is equivalent to an average probability value from Table 2. The low and high likelihood thresholds are logically derived on the basis of this value (i.e. a new mode is calculated if the customer type's likelihood to execute a particular decision is not moderate). The same method is used for adapting an average delay value from Table 1.

A key aspect to consider is that the most interesting system outcomes evolve over time and many of the goals of the retail company (e.g., service standards) are planned strategically over the long-term. In ManPraSim v2, we have introduced a finite population where each agent is given a certain characteristic based on one out of five possible types mentioned above. Once agents are created, they are added to a customer pool. Each hour a certain amount of agents chosen at random from the agents in the customer pool are released into the department at an exponential rate based on the footfall value for that hour. When the agent has finished shopping, statistics are updated (amongst them the customer satisfaction index value) and the agent returns to the customer pool. A customer retains his or her customer satisfaction index throughout the runtime of the simulation.

We have also added some transitions that allow the emulation of customers' behaviour when the branch is closing. These are immediate exits of a state that are triggered when the shop is about to close. Not all states have these additional transitions because it is for example very unlikely that a customer leaves the branch immediately when he/she is already queuing to pay. Now the simulated department empties within a ten to fifteen minute period, which conforms to what we observed in the real system.

*3.4.3 Testing the model*
In order to test the operation of ManPraSim v2 and ascertain its face validity we conducted several preliminary experiments. It turned out that conducting the experiments with the data we collected during our case study did not provide us with a satisfactory match to the performance data of the real system. We identified the staffing setup used in the simulation model as the root cause of the problem.

The data we have used here has been derived from real staff rotas. On paper, these real rotas suggested that all staff are engaged in exactly the same work throughout the day but we knew from working with and observing staff in the case study organisation that in reality each role includes a variety of activities. Staff members in the real organisation allocate their time between competing tasks such as customer service, stock replenishment and taking money. Our simulation model incorporates only one type of work per staff member. For example, the A&TV staff rota indicates that only one dedicated cashier works on weekdays. When we have attempted to model this arrangement, customer queues have become extremely long, and the majority of customers ended up losing their patience and leaving the department prematurely with a high level of dissatisfaction. In the real system, we observed other staff members working flexibly to meet the customer demand, and if the queue of customers grew beyond a certain point then one or two would step in and open up further tills to take customers' money before they became dissatisfied with waiting. Furthermore, we observed that a service staff member, when advising a customer, would often continue to close the sale (e.g. filling in guarantee forms and taking the money off the customer) rather than asking the customer to queue at the till and moving on to the next customer.

This means that currently our abstraction level is too high and we do not model the real system in an appropriate way. In our experiments we have modulated the staffing levels to allow us to observe the effects of changing key variables but we have tried to maintain the main characteristic differences between the departments (i.e. we still use more staff in the WW department compared to the A&TV department, only the amount has changed). We hope to be able to fix this problem in a later version of our simulation model. For now, we do not consider this a big problem as long as we are aware of it.

**4 EXPERIMENTS**
As mentioned earlier our simulation model has been developed in two major steps. After finishing the first step, we have conducted a set of experiments to investigate the usefulness of the developed simulation model for studying the impact of people management practices on productivity and customer satisfaction (Siebers *et al*, 2007a; 2007b). We investigated the following scenarios:

- ***Branch operation:*** (1) Varying the number of open tills and consequently the mix of staff roles (as we kept the overall staffing level constant) to assess the effect on department performance. (2) Testing the impact of expert staff availability on customer satisfaction.
- ***Empowerment:*** (1) Varying the extent to which a cashier can independently decide whether or not to make a customer refund. (2) Testing the impact of non-expert staff members choosing whether or not to stay with his or her customer when the customer requires advice from an expert. If they choose to stay, the original staff member can learn from the interaction.
- ***Training:*** Mimicking an evolutionary process, whereby normal staff members can progressively develop their product knowledge over a period of time and at an agreed level of competence will be promoted to expert status.

With ManPraSimMod v2 we have conducted a validation experiment to test our customer pool implementation and we have conducted a sensitivity analysis to investigate the impact of our customer types on simulation output. These experiments are not conducted to provide insight into the operations of the case study department. Instead, they are carried out to test the simulation model behaviour.

**4.1 TESTING THE CUSTOMER POOL IMPLEMENTATION**
For testing the customer pool implementation of ManPraSim v2 we have repeated the first of the above experiments with our latest model. Our reasoning behind this experiment is that a correct implementation should give us similar results as we obtained previously with the older version, if we use an even mix of all five customer types available.

*4.1.1 Experiment description*
Our case study work had helped us to identify some distinguishing characteristics of the two department types under study (e.g. customer arrival rates and customer service times). In the experiment we examined the impact of these individual characteristics on the number of transactions and two customer satisfaction indices. First, the number of satisfied customers (how many customers left the store with a positive service level index value) and second the overall satisfaction level (the sum of all customers' service level index values). During the experiment, we held the overall number of staffing resources constant at ten, the simulation lifespan at ten weeks and we conducted 20 replications for each model configuration to enable the application of rigorous statistical techniques. We focussed on the composition of the team's skills by systematically varying the proportion of staff allocated to each role within the pool of ten staff. In each department, staff was allocated to either selling or cashier duties. In reality, we saw that allocating extra cashiers would reduce the shop floor sales team numbers, and therefore the total number of customer-facing staff in each department is kept constant.

When we conducted the experiment with ManPraSim v1, we found support largely in favour of the predicted curvilinear relationship between the number of cashiers and each of the outcome variables. We expected this type of relationship because limiting factors restrict the more extreme experimental conditions; very small numbers of cashiers available to process purchase transactions detrimentally impacted on the volume of customer transactions, and conversely very small numbers of selling staff restricted opportunities for customers to receive advice and consequently negatively influenced customer perceptions of satisfaction. We had also predicted that performance outcomes would peak with a smaller number of cashiers in A&TV as compared to WW given the greater customer service requirement in A&TV, and the higher frequency of sales transactions in WW. Results supported this hypothesis for both customer satisfaction indices, but not for the number of transactions where the peak level occurred at the same point. This was surprising because we would have expected the longer average service times in A&TV to put a greater 'squeeze' on customer help availability with even a relatively small increase in the number of cashiers.

When repeating the experiment with ManPraSim v2 we tried to mimic the generic customer from v1 by using an even mix of all five customer types available. We maintained a customer pool size at 10,000 for each of the model configurations. In order to enable the application of rigorous statistical techniques we have conducted 20 replications. The experimental results are analysed using tabulated and graphical illustrations of mean values. Despite our prior knowledge of how the real system operates, we were unable to hypothesize precise differences in variable relationships. We have instead predicted patterns of relationships and we believe this is congruent with what the simulation model can offer; it is a decision-support tool which is only able to inform us about directional changes between variables (actual figures are notional).

*4.1.2 Hypotheses*
In general, we predict a similar number of transactions for both simulation model versions as we tried to mimic a generic customer. We do however predict a change in the number of satisfied customers: we expect the results to show a shift from neutral to either satisfied or dissatisfied. This polarisation of customer satisfaction is expected because ManPraSim v2 enables the customer population to re-enter the system and each re-entry increases the likelihood that neutral customers shift to satisfied or dissatisfied. Looking at overall satisfaction level, we would expect similar trends for v1 and v2, but we predict the magnitude of the results for v2 to be significantly higher. This is because it incorporates an accumulated history of satisfaction trends for customers who have returned to the department on multiple occasions, unlike v1, which records satisfaction levels only for single, independent visits.

*4.1.3 Results*
The descriptive statistics for ManPraSim v1 and v2 are shown in Table 4 and graphical representations of the results are presented in Figure 4.

| Dept | Staffing No. cashiers | ManPraSim v1 | | | | | | ManPraSim v2 | | | | | |
|---|---|---|---|---|---|---|---|---|---|---|---|---|---|
| | | No. transactions | | No. satisfied customers | | Overall satisfaction level | | No. transactions | | No. satisfied customers | | Overall satisfaction level | |
| | | Mean | SD | Mean | SD | Mean | SD | Mean | SD | Mean | SD | Mean | SD |
| A&TV | 1 | 4864.10 | 23.93 | 12341.50 | 74.03 | 6476.40 | 599.10 | 6124.80 | 16.23 | 15528.40 | 242.63 | 2624.70 | 3,110.39 |
| | 2 | 9,811.65 | 49.29 | 14826.30 | 73.63 | 17,420.35 | 513.50 | 11857.90 | 38.99 | 19226.90 | 241.13 | 25803.90 | 2,740.35 |
| | 3 | 14481.50 | 79.05 | 17,656.65 | 73.01 | 26470.10 | 571.53 | 15,415.85 | 82.41 | 22,150.00 | 237.35 | 45,162.45 | 3,571.49 |
| | 4 | 15,064.55 | 90.21 | 17683.90 | 101.85 | 30,997.05 | 502.73 | 15,374.25 | 102.91 | 21635.80 | 197.24 | 50341.10 | 2,828.70 |
| | 5 | 14,190.75 | 73.20 | 16494.20 | 73.24 | 26,746.35 | 701.66 | 14,538.85 | 82.20 | 19,953.85 | 219.76 | 36184.60 | 2,219.08 |
| | 6 | 13302.30 | 121.30 | 15,315.35 | 116.97 | 18274.10 | 546.14 | 13,622.65 | 92.50 | 17,979.55 | 134.00 | 12,407.05 | 1,942.33 |
| | 7 | 12338.50 | 112.38 | 13997.40 | 94.97 | 7444.30 | 569.90 | 12586.50 | 110.15 | 15,956.25 | 165.08 | -14449.10 | 2,044.52 |
| WW | 1 | 8053.20 | 30.74 | 18310.70 | 115.18 | 20631.20 | 694.64 | 10235.60 | 24.44 | 28,532.95 | 273.29 | 51820.40 | 2,583.76 |
| | 2 | 15737.80 | 64.44 | 22427.30 | 103.34 | 45671.40 | 700.99 | 20,051.25 | 26.43 | 38,279.05 | 188.83 | 146,838.95 | 2,968.69 |
| | 3 | 25,602.45 | 101.89 | 28743.70 | 91.52 | 62,239.15 | 546.85 | 27987.20 | 113.67 | 47,663.95 | 229.61 | 229,416.15 | 2,976.82 |
| | 4 | 29,154.95 | 193.98 | 30,999.05 | 190.91 | 77,333.95 | 663.16 | 28985.50 | 154.59 | 48,262.15 | 239.31 | 285100.10 | 5,081.91 |
| | 5 | 27893.20 | 156.54 | 29455.90 | 157.78 | 77827.30 | 470.96 | 27887.25 | 165.11 | 45783.10 | 264.10 | 275,291.05 | 3,236.96 |
| | 6 | 26,189.55 | 88.56 | 27,536.25 | 93.50 | 67,958.65 | 632.84 | 26,488.45 | 173.30 | 42,485.95 | 277.40 | 230957.30 | 3,955.71 |
| | 7 | 24496.70 | 183.85 | 25634.50 | 184.57 | 53262.90 | 801.20 | 24,632.85 | 100.57 | 38,252.45 | 300.09 | 169577.80 | 3,517.38 |

**Table 4:** Descriptive statistics for experiment 1 (all to 2 d.p.).

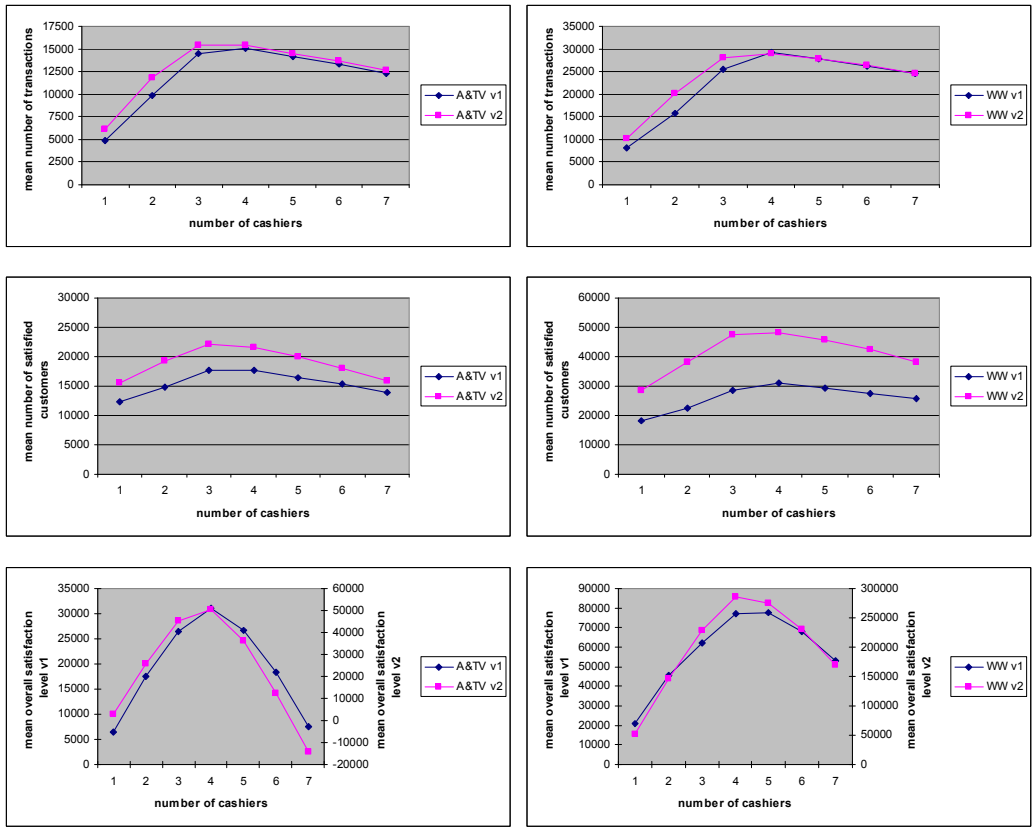

**Figure 4:** Diagrams for experiment 1.

Looking at the number of transactions for both departments, it is clear that both simulation model versions produce a highly similar pattern of results. The number of satisfied customers is higher across all conditions of both departments for ManPraSim v2. This is as predicted and interestingly very high levels of satisfaction can be seen in WW in particular. We attribute this to the higher transaction volumes in WW coupled with our expectations of v2 resulting in higher levels of customer satisfaction as customers visit the branch on multiple occasions and commit to polarised opinions. Examining the overall satisfaction level, our hypotheses hold; results for both departments clearly follow the same trends regardless of simulation model version. In summary, all results are as predicted.

*4.1.4 Required model improvement*

With the introduction of a finite population (represented by our customer pool), we had to rethink the way in which we collect statistics about the satisfaction of customers. Previously, the life span of a customer has been a single visit to the department. At the end of his or her visit, the individual's satisfaction score (direction and value) has been recorded. Now the life span of a customer lasts the full runtime of the simulation and he or she can be picked several times to visit the branch during that period. Our previous customer satisfaction measures now collect some different information: satisfaction scores considering customers' cumulative satisfaction history. These measures do not reflect individuals' satisfaction with the current service experience but instead the satisfaction with the overall service experience during the lifetime of the agent. Furthermore, they are biased to some extent in that an indifferent rating quickly shifts into satisfaction or dissatisfaction (arguably, this is realistic, because most people like to make a judgement one way or the other). Whilst still a valuable piece of information we would also like to know how the current service is perceived by each customer. For this reason, we have introduced a set of new performance measures to record the experience of each customer's individual visit. These are the same measures as before but they ignore the customer's previous experiences. They have been added to ManPraSim v2 before running the next experiment.

**4.2 TESTING THE CUSTOMER TYPE IMPLEMENTATION**

The main purpose of this experiment is to test the sensitivity of the simulation results toward our new defined customer types. In addition, we use this experiment to test the robustness of our new experience per visit measures. This experiment should demonstrate how useful these new measures are.

*4.2.1 Experiment description*

The departmental managers reported that they find mainly solution demanders and service seekers in the A&TV department while the WW department is mainly visited by shopping enthusiasts. We have used this real customer split configuration amongst other variations to configure the customer population in our second experiment. In total seven customer type configurations are tested for both department types. The composition of each configuration is given in Table 5. The first five configurations (a-e) are extreme customer type settings. These occur when 100% of the customer population behaves according to the same customer type. Extreme configurations amplify the impact of differences between the behaviour of different customer types. For the final two configurations, (f) uses an equal composition of each customer type and (g) uses a real customer split reflecting that reported by managers working in the case study departments.

| configuration → <br> customer stereotype ↓ | a | b | c | d | e | f | g <br> (A&TV) | g <br> (WW) |
|---|---|---|---|---|---|---|---|---|
| shopping enthusiasts | 10000 | 0 | 0 | 0 | 0 | 2000 | 500 | 8000 |
| solution demanders | 0 | 10000 | 0 | 0 | 0 | 2000 | 4000 | 0 |
| service seekers | 0 | 0 | 10000 | 0 | 0 | 2000 | 4000 | 0 |
| disinterested shoppers | 0 | 0 | 0 | 10000 | 0 | 2000 | 500 | 2000 |
| internet shoppers | 0 | 0 | 0 | 0 | 10000 | 2000 | 1000 | 0 |

**Table 5:** Definition of customer type configurations.

The customer pool size is maintained at 10,000 for each model configuration, and 20 replications are conducted for every model configuration to enable the application of rigorous statistical techniques. As we explained previously, we test hypotheses asserting directional relationships.

*4.2.2 Hypotheses*

For our sensitivity analysis, we are particularly interested in drawing comparisons between department types and the extreme customer type configurations. The latter two configurations (f) and (g) are still of interest however these are composite configurations, resulting in a less significant or 'dumped down' effect on department performance measures.

We predict that a greater number of customers leave satisfied following a transaction in WW than in A&TV because of the higher frequency of transactions in WW. We also expect that (d) and (e) experience relatively low counts, across both departments, due to the low likelihood that either of these customer types makes a sales transaction.

We hypothesise that more customers leave before receiving normal help in A&TV due to the longer average service times than in WW, with the exception of (b) and (d) which both have a low likelihood of customers requesting help and are therefore linked to extremely low premature customer departures across both A&TV and WW. We predict that configurations (c) and (e) result in the highest counts on this measure, across both departments, because of the high service demand placed on normal staff in this department by these two customer types.

Again, we expect more customers to leave before receiving expert help in A&TV due to longer service times. This time we do predict a difference for (b) and (d) in this direction given the importance of expert advice to A&TV customers, but we expect this difference to remain smaller than under different customer type configurations due to the relatively low propensities of these customer types to seek advice. Again, we hypothesise that (c) and (e) result in the highest counts of premature customer departures because these customer types in particular demand a great deal of advice.

We predict that a significantly greater number of customers leave whilst waiting to pay in A&TV than in WW across customer types because of the longer average cashier service time, resulting in longer queues at the till and therefore more customers leaving prematurely whilst waiting to pay. We hypothesise that configurations (a) and (b) are linked to higher numbers of customers leaving before paying because these customer types have a high likelihood of making a purchase. Conversely, we hypothesise that (d) and (e) are linked to lower numbers of customers leaving before paying because these customer types have a lower likelihood of making a purchase.

We hypothesise that the absolute number of customers who leave without finding anything is greater in WW than in A&TV across customer types. This is because even though the conversion rate in WW is slightly higher, the footfall is much greater in WW (i.e. customers visit more frequently). We predict that the greatest counts of customers leaving without a purchase are for (d) and (e) due to the low likelihood of these customer types making a purchase. We expect that the lowest counts of customers leaving without a purchase are for (a) and (b) because these customer types have a high propensity for retail purchases.

For customer satisfaction indices, we predict that the measure which allows customers to remember their past visits and to accumulate an overall satisfaction score, results in more pronounced effects than those using the other 'score per visit' measure. We predict that for A&TV (c) and (e) are linked to a relatively high proportion of dissatisfied customers, because these configurations place the greatest service demands on staff and therefore under these extreme conditions staff cannot always satisfactorily meet the demand for advice. This effect is much clearer for the ongoing satisfaction scores because customers remember past dissatisfactory experiences. We predict that (d) result in an extremely low percentage of dissatisfied customers because it is unlikely that the low demands of this disinterested customer type do not stretch the staffing constraints to the point where they cannot be met. To these hypotheses further, we predict that where customer's current customer satisfaction is anchored by their previous perceptions of satisfaction; we expect a smaller proportion of neutral customer satisfaction scores (compared to the experience per visit measure). This occurs because over time customers will accumulate more experience of the department and take on more polarised opinions.

*4.2.3 Results*
A series of two-way between groups ANOVAS have been used to assess the impact of customer types on counts of customer leaving satisfied after a transaction with a cashier, whilst waiting for normal or expert help, whilst waiting to pay, or leaving without finding a suitable purchase. The descriptive statistics for the experiment are shown in Table 6, followed by graphs for each performance variable. Where tests indicate significant differences, Tukey's post hoc tests have been applied to ascertain exactly where these differences occur for different customer type configurations. Stacked bar charts have been examined to assess the impact of customer type configuration on two customer satisfaction indices: considering the customer's previous experiences and considering the customer's experience per visit.

| Dept. | Customers ... Configuration | ... leaving happy | | ... leaving before normal help | | ... leaving before expert help | | ... before paying | | ... without finding anything | |
|---|---|---|---|---|---|---|---|---|---|---|---|
| | | Mean | SD | Mean | SD | Mean | SD | Mean | SD | Mean | SD |
| A&TV | (a) | 12,231.70 | 24.42 | 981.95 | 66.28 | 450.50 | 18.77 | 12,339.65 | 121.14 | 14,863.40 | 147.70 |
| | (b) | 12,069.30 | 27.36 | 9.05 | 4.15 | 112.05 | 11.44 | 11,892.30 | 160.10 | 16,747.35 | 121.65 |
| | (c) | 11,906.10 | 44.41 | 6,457.75 | 134.34 | 569.50 | 25.47 | 3,285.95 | 88.61 | 18,621.80 | 117.60 |
| | (d) | 8,271.55 | 74.36 | 10.70 | 2.58 | 110.75 | 9.45 | 293.25 | 29.75 | 32,151.00 | 171.51 |
| | (e) | 8,417.00 | 56.92 | 6,487.30 | 151.49 | 557.95 | 23.36 | 129.95 | 12.10 | 25,301.25 | 129.65 |
| | (f) | 11,881.50 | 38.01 | 1,002.00 | 62.62 | 439.65 | 15.80 | 5,010.45 | 100.73 | 22,581.35 | 162.87 |
| | (g) | 12,055.70 | 32.19 | 1,125.05 | 59.93 | 457.85 | 19.81 | 7,091.40 | 170.28 | 20,191.80 | 152.39 |
| WW | (a) | 30,036.90 | 53.58 | 12.40 | 6.24 | 108.60 | 10.00 | 11,107.40 | 232.03 | 22,548.95 | 175.17 |
| | (b) | 29,753.80 | 44.79 | 0.05 | 0.22 | 22.65 | 4.87 | 9,336.50 | 194.48 | 24,718.45 | 128.15 |
| | (c) | 29,311.90 | 57.11 | 495.15 | 45.92 | 252.60 | 18.62 | 3,060.70 | 95.13 | 30,631.35 | 161.12 |
| | (d) | 15,664.85 | 98.84 | 0.00 | 0.00 | 19.85 | 5.15 | 1.75 | 2.10 | 48,131.15 | 228.92 |
| | (e) | 22,075.20 | 166.99 | 521.30 | 44.89 | 248.00 | 18.18 | 16.20 | 5.26 | 41,022.60 | 210.83 |
| | (f) | 28,247.55 | 101.86 | 11.40 | 4.42 | 104.60 | 12.96 | 1,908.45 | 91.21 | 33,500.85 | 238.00 |
| | (g) | 29,666.25 | 30.28 | 4.50 | 3.27 | 80.05 | 10.45 | 6,320.85 | 204.21 | 27,664.90 | 195.88 |

**Table 6:** Descriptive Statistics for ANOVA Variables (all to 2 d.p.).

Levene's test for equality of variances was violated ($p<.05$) for customers leaving satisfied, before receiving normal or expert help, and whilst waiting to pay. To address this, ANOVAs investigating these variables used a more stringent significance level ($p<.01$).

For customers leaving satisfied, there were significant main effects for both department [$F(1, 266) = 3,251,075$, $p<.001$] and customer type configuration [$F(6, 266) = 101,910.1$, $p<.001$], with a significant interaction effect [$F(6, 266) = 28,651.93$, $p<.001$]. Post hoc tests revealed significant differences for every single comparison ($p<.001$) apart from between (b) and (g) ($p=.029$). Looking at Figure 5a, there is undisputed support for all of the hypotheses.

Comparing the count of customers leaving before receiving normal help, significant effects occurred for both department [$F(1, 266) = 79,090.42$, $p<.001$] and customer type configuration [$F(6, 266) = 24,058.12$, $p<.001$], plus a significant interaction effect [$F(6, 266) = 17,180.46$, $p<.001$]. Tukey's tests demonstrated significant differences for all comparisons ($p<.001$) apart from three pairings: (a) and (f), (b) and (d), and (c) and (e). All of the predictions have been borne out (see Figure 5b). In particular, the significantly higher count of customers leaving prematurely for (c) and (e) is very pronounced.

Results for customers leaving before receiving expert help revealed significant effects for both department [$F(1, 266) = 19,733.82$, $p<.001$] and customer type configuration [$F(6, 266) = 3,147.72$, $p<.001$], plus a significant interaction effect [$F(6, 266) = 593.71$, $p<.001$]. Post hoc tests demonstrated significant differences between all but four comparisons ($p<.001$). Looking at Figure 5c, the pattern of results is as hypothesized.

For customers leaving whilst waiting to pay, there were significant effects for both department [$F(1, 266) = 5,750.20$, $p<.001$] and customer type configuration [$F(6, 266) = 51,939.13$, $p<.001$], with a significant interaction effect [$F(6, 266) = 838.79$, $p<.001$]. Tukey's tests revealed significant differences for all comparisons ($p<.001$) apart from between (d) and (e). Results are presented in Figure 5d and display consistent support for the hypotheses.

Results for customers leaving before finding anything to buy revealed significant effects for both department [$F(1, 266) = 293,989.90$, $p<.001$] and customer type configuration [$F(6, 266) = 75,977.11$, $p<.001$], plus a significant interaction effect [$F(6, 266) = 4,573.90$, $p<.001$]. Post hoc tests revealed significant differences between all comparisons ($p<.001$). The pattern of results is as hypothesized (see Figure 5e).

To investigate the impact of customer types on satisfaction indices, the mean customer satisfaction ratings have been calculated for each configuration and have been displayed in 100% stacked bar charts. Figure 6 shows customer satisfaction considering the customer's previous experiences, and Figure 7 shows customer satisfaction considering the customer's experience per visit. There is much differentiation in satisfaction scores across the contrasting customer type configurations. Evidence supports all hypotheses.

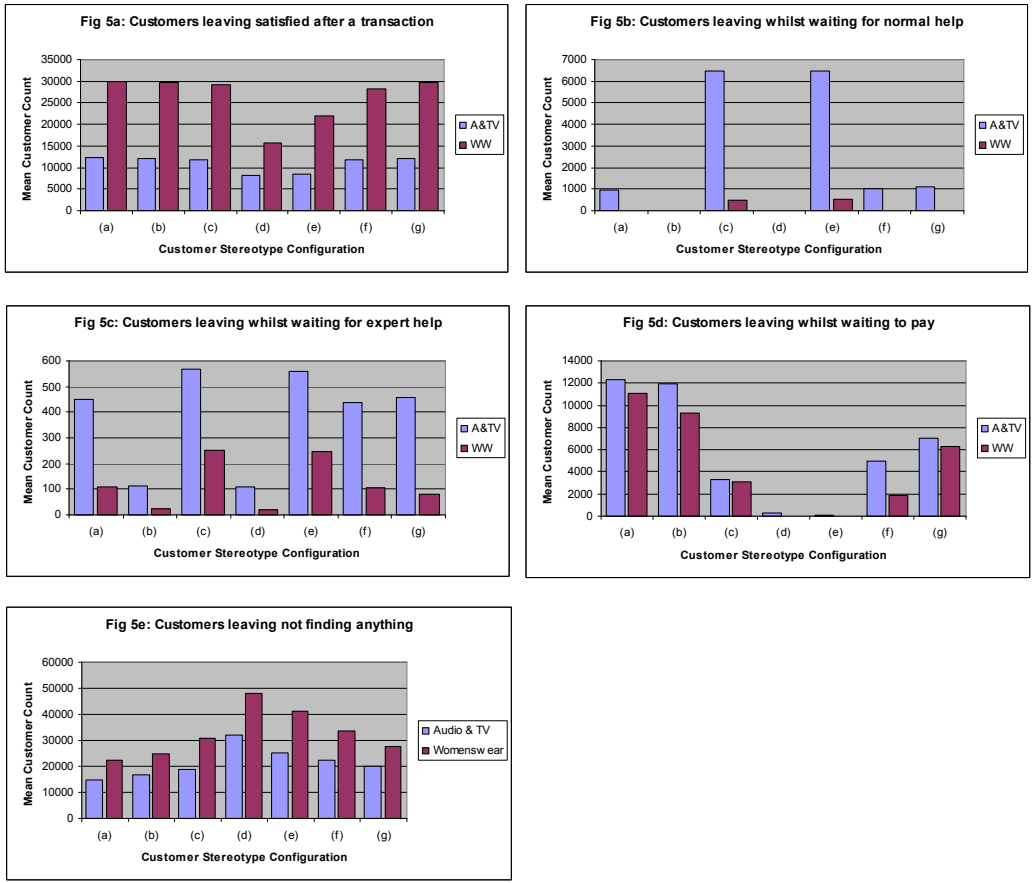

**Figure 5:** Results from experiment 2 - Customer leaving state split.

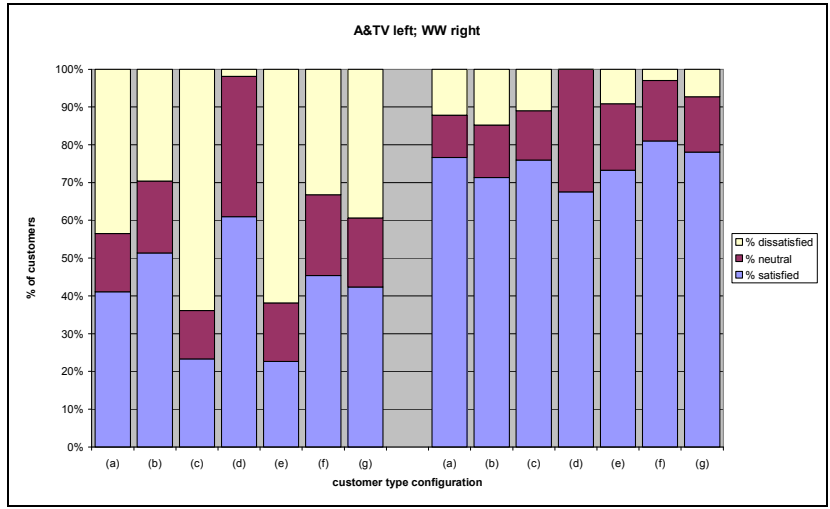

**Figure 6:** Results from experiment 2 - customer satisfaction (considering history, in % of total customers).

Overall, our results indicate that customer types exert a considerable impact on system performance as demonstrated by a number of complementary measures. In addition, we have shown that our new customer satisfaction measures are a useful asset for analysing the service quality as perceived by the customer at each individual visit. We have presented evidence demonstrating that our improved customer satisfaction measure produces results closer to what we would expect from the real system, and the increased polarisation of customers' satisfaction is a further reason to select carefully which customer types to implement in the simulation model.

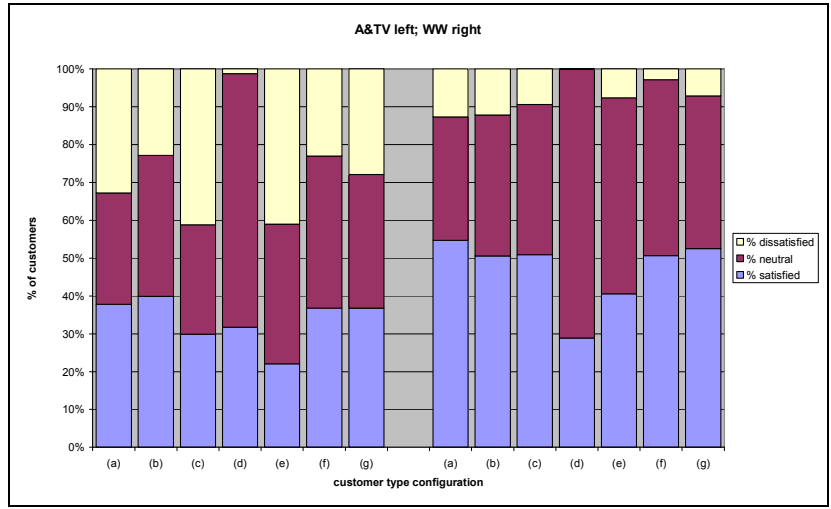

**Figure 7:** Results from experiment 2 - customer satisfaction (per visit, in % of total visits).

## 5 CONCLUSIONS

In this paper, we have presented an alternative to the modelling approaches commonly used for investigating the impact of management practices on retail performance. Simulation modelling has proven to be a very useful tool for this kind of analysis. In particular, the combination of DEM for representing the system and ABM for representing the entities inside the system seems to be the formula to success for building simulation models of dynamic heterogeneous people-centric systems in an OR context.

Our investigations for this paper have focused on the question of what level of abstraction to use for this kind of simulation models. We have added several features to our initial functional representation of the real system. Our latest version now includes more complex operational features to make it a more realistic representation closer to the real retail environment that we observed and also we have developed our agents further, enhancing their heterogeneity and decision making capabilities. We have tested some of these new features here by conducting validation experiments as well as a sensitivity analysis. However, more tests are required to establish the usefulness of all the enhancements implemented in the latest simulation model.

After the first experiment, we realised that with the introduction of a finite population we also had to rethink the way in which we collect statistics about the satisfaction of customers. The existing measures did not reflect individuals' satisfaction with the current service experience any more but instead measured the satisfaction with the overall service experience during the lifetime of the agent. We have added some customer satisfaction measure to evaluate how the current service is perceived by each customer. These measures have shown to provide some useful information for analysing the service quality as perceived by the customer during each visit and results are much closer to what we would expect from the real system.

A big advance has been the implementation of a diverse population by introducing customer types. These allow a better representation of the different service needs that customers have in different departments and the different responses to the service provided which is also apparent in the real world customer population. Through introducing customer types, we have been able to better define the heterogeneous customer groups who visit the different case study departments and observe previously hidden differences in the impact of the people management practices on the different departments. In our second experiment, we have demonstrated that customer types exert a considerable impact on system performance. It is therefore important that practitioners invest time and effort in analysing the existing types of customers and their actual proportions and derive an effective way of characterising the differences between these groups.

We have used likelihoods to buy, wait and request help for defining our customer types. However, there are many more categories one could use to distinguish different customer types, e.g. how frequent customers come back or how many items they buy per visit. Much of the required data could be available from loyalty cards or are collected by the marketing department of the company. Building on this advance is the introduction of a finite population with an enduring memory of customers' previous shopping experiences for each individual agent. This allows the agents to change their behaviour (e.g. their patience level) according to their previous experiences. These two enhancements enable us for the first time to study long-term effects of managerial decisions on specific customer groups with our simulation model, which opens up a new range of problem scenarios that can be investigated. This is particularly important as many managerial decisions are developed for the long-term (to encourage trust and loyalty from both staff and customers) and are unlikely to demonstrate their full potential after just a single visit of an individual customer.

Overall, we can affirm that the upgrades we have introduced and tested so far are all useful assets to the simulation model. They allow us to obtain a broader understanding of the situation and to investigate many issues and questions we could not previously investigate as for example long-term effects of people management practices on specific customer groups.

A major limitation in ManPraSim v2 is the absence of consideration of staff proactiveness. What we have actually observed during our case study and what is also encouraged by the guidelines for staff of the case study company is that employees act proactively, for example by approaching customers within a set period of time, or opening tills when queues grow beyond a defined level. In addition, if we want to use real staffing data in order to enhance the predictive capabilities of our simulation model relative to the real system, we need to model how staff allocate their tasks between competing activities rather than focusing on one type of work. Considering these features will allow us to increase the grade of realism to match the behaviour of the actors in the model better to those in the real system. This will also allow us to achieve a better match when comparing the system performance measures (e.g. transactions or staff utilisation) between our simulation model and the real system. Luckily, ABM is a technique that supports the modelling of proactive agent behaviour. First however, we need to test our remaining enhancements by conducting more sensitivity analyses, to see if these are useful assets or not.

Once these limitations have been eradicated, we would like to conduct some more fundamental investigations. As we stated earlier the most interesting system outcomes evolve over time and many of the goals of a retail business (e.g. service standards) form part of a long-term strategy. It would be interesting to see under which circumstances the demand for services varies in an unplanned fashion and how well staff can cope with it. A common application area for ABM is modelling the diffusion of innovations in social networks (Bonabeau, 2002; Garcia, 2005; Janssen and Jager, 2002). We would like to use the ideas developed here and transfer them into a service context to model diffusion of customer experiences in form of word-of mouth networks. With our enhanced agents, we are now in the position to investigate this kind of issues.

Finally, we are interested in exploring other ways of implementing the agent decision making process. It has been argued that modelling the autonomous internal decision making logic of customers is a crucial element for simulation models of consumer behaviour (Jager, 2007). It would be interesting to compare such an approach to the probabilistic one we have currently in place, in particular as no such study has been found in the literature.

In conclusion, we can say that the multidisciplinary of our team has helped us to gain new insight into the behaviour of staff and customers retail organisations. The main benefit from adopting this approach is improved understanding of, and debate about, a problem domain. The very nature of the methods involved forces researchers to be explicit about the rules underlying behaviour and to think in new ways about them. As a result, we have brought work psychology and simulation modelling together to form a new and exciting research area.